# Exploiting the potential of unlabeled endoscopic video data with self-supervised learning

Tobias Ross · David Zimmerer · Anant Vemuri ·
Fabian Isensee · Manuel Wiesenfarth · Sebastian
Bodenstedt · Fabian Both · Philip Kessler ·
Martin Wagner · Beat Müller · Hannes Kenngott ·
Stefanie Speidel · Annette Kopp-Schneider · Klaus
Maier-Hein · Lena Maier-Hein

**Abstract**

*Purpose* Surgical data science is a new research field that aims to observe all aspects of the patient treatment process in order to provide the right assistance at the right time. Due to the breakthrough successes of deep learning-based solutions for automatic image annotation, the availability of reference annotations for algorithm training is becoming a major bottleneck in the field. The purpose of this paper was to investigate the concept of self-supervised learning to address this issue.

*Methods* Our approach is guided by the hypothesis that unlabeled video data can be used to learn a representation of the target domain that boosts the performance of state-of-the-art machine learning algorithms when used for pre-training. Core of the method is an auxiliary task based on raw endoscopic video data of the target domain that is used to initialize the convolutional neural network (CNN) for the target task. In this paper, we propose the re-colorization of medical images with a generative adversarial network (GAN)-based archi-

Tobias Ross[1] · Anant Vemuri[1] · Lena Maier-Hein[1] · David Zimmerer[2] · Fabian Isensee[2] · Klaus Maier-Hein[2] ·
Manuel Wiesenfarth[3] · Annette Kopp-Schneider[3]
E-mail: t.ross@dkfz-heidelberg.de
[1]Computer Assisted Medical Interventions, [2]Medical Image Computing, [3]Division of Biostatistics
German Cancer Research Center
Im Neuenheimer Feld 581
69210 Heidelberg, Germany

Sebastian Bodenstedt · Stefanie Speidel
Translational Surgical Oncology
National Center for Tumor Diseases (NCT)
Fetscherstrasse 74
01307 Dresden, Germany

Marting Wagner · Beat Müller · Hannes Kenngott
Deparment of General, Visceral and Transpl. Surgery
University of Heidelberg
Im Neuenheimer Feld 110
69210 Heidelberg, Germany

Fabian Both · Philip Kessler
understand.ai
Hirschstr. 71
76133 Karlsruhe, Germany

tecture as auxiliary task. A variant of the method involves a second pre-training step based on labeled data for the target task from a related domain. We validate both variants using medical instrument segmentation as target task.

*Results*   The proposed approach can be used to radically reduce the manual annotation effort involved in training CNNs. Compared to the baseline approach of generating annotated data from scratch, our method decreases exploratively the number of labeled images by up to 75% without sacrificing performance. Our method also outperforms alternative methods for CNN pre-training, such as pre-training on publicly available non-medical (COCO) or medical data (MICCAI EndoVis2017 challenge) using the target task (in this instance: segmentation).

*Conclusions*   As it makes efficient use of available (non-)public and (un-)labeled data, the approach has the potential to become a valuable tool for CNN (pre-)training.

# 1 Introduction

Surgical data science is an evolving research field that aims to "observe all that is occurring within and around the treatment process" in order to "improve the quality of interventional healthcare and its value by capturing, organizing, analyzing and modelling data" [17]. An international consortium comprising leading researchers from engineering and medicine suggested that context-aware assistance in minimally-invasive surgery may be a key clinical application of surgical data science [17]. The computer vision challenges in this context include detection, segmentation and tracking of medical devices in endoscopic video data, organ classification, and surgical action/phase recognition. While extremely promising results can be obtained with state-of-the-art supervised machine learning approaches, typically, the methods do not generalize well. An example is provided in Fig. 1, in which a state-of-the-art convolutional neural network (CNN) performs well when trained and tested on the data from the MICCAI instrument tracking challenge 2017[1], organized as part of the MICCAI endoscopic vision challenge 2017[2]. However, mean performance drops by more than 50% when applied to endoscopic video data from another site. This is an important limitation as curation of (sufficient) training data is extremely labor-intensive and is currently hindering progress in the field. Related methods address this challenge with crowdsourcing-based approaches [15, 16], i.e., methods which outsource annotation tasks to masses of anonymous workers in an online community. In this paper, we investigate an entirely new approach that has been inspired by recent achievements in the field of self-supervised learning (see e.g. [1, 4, 22, 23, 30]) and is based on the observation that it is often the small amount of *annotated* medical image data rather than the amount of raw medical data that causes the bottleneck related to training data acquisition in surgical data science. Our hypothesis is that masses of unlabeled video data can be used to learn a representation of the target domain that can boost the performance of state-of-the-art machine learning algorithms when used for pre-training. We investigate this hypothesis by using the CNN-based medical instrument segmentation as an example. Sec. 2 presents the general concept, a first prototype implementation and the study design for hypothesis validation. The results are presented in sec. 3 followed by a discussion of our findings in the context of related work in sec 4.

---

[1] https://endovissub2017-roboticinstrumentsegmentation.grand-challenge.org/
[2] https://grand-challenge.org/site/endovis/endoscopic_vision_challenge/

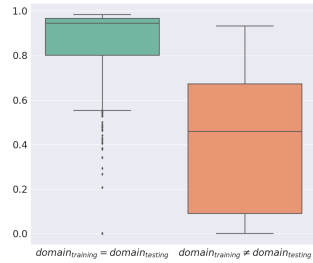

Fig. 1: Limited generalization capabilities of state-of-the-art machine learning algorithms. When training and testing are performed on data from the same hospital, state-of-the-art segmentation performance is achieved with the algorithm presented in sec. 2.2. However, when the same model is applied to data from a different site, accuracy drops dramatically.

## 2 Methods

### 2.1 Concept overview

Our approach, which we refer to as *Pre-training with Auxiliary Task (PAT)*, is illustrated in Fig. 2 and has the following components:

- **Target task**: Endoscopic vision task to be solved by the algorithm, e.g. segmentation of medical instruments from endoscopic video data
- **Unlabeled data**: Large number $N_{unlabeled}$ of unlabeled endoscopic images $I_{unlabeled}$ that are representative of the target domain (e.g. laparosopic video data from a specific hospital for a specific application)
- **Labeled data**: Comparatively small number $N_{labeled} << N_{unlabeled}$ of images $I_{labeled}$ labeled according to the target task.
- **Architecture for target task**: A CNN-based architecture designed to solve the target task using $I_{labeled}$, e.g. a U-Net [25]
- **Auxiliary task**: Task designed to leverage information in the unlabeled data (e.g. image re-colorization as described below).
- **Architecture for auxiliary task**: A CNN-based architecture designed to solve the target task with a self-supervised learning approach using $I_{unlabeled}$.

The core of the method is the auxiliary task which leverages the information available in unlabeled image data from the target domain for to improve the generalization capabilities of CNNs. In this paper, we use an adversarial approach to train the target-task network to re-colorize grayscale images. Labeled training data is then used to refine the model for the task of interest (here: segmentation). The following section gives a concrete example on how to instantiate the concept.

### 2.2 Prototype implementation

We implemented the concept proposed using re-colorization as auxiliary tasks (sec. 2.2.1) and medical instrument segmentation (for which CNNs are currently the most widespread used method [5,6,21,29]) as target task (sec. 2.2.2). Important design decisions are the usage of a combined reconstruction and adversarial loss for realistic re-colorization results and a U-Net which is a commonly used target-task network for the segmentation task.

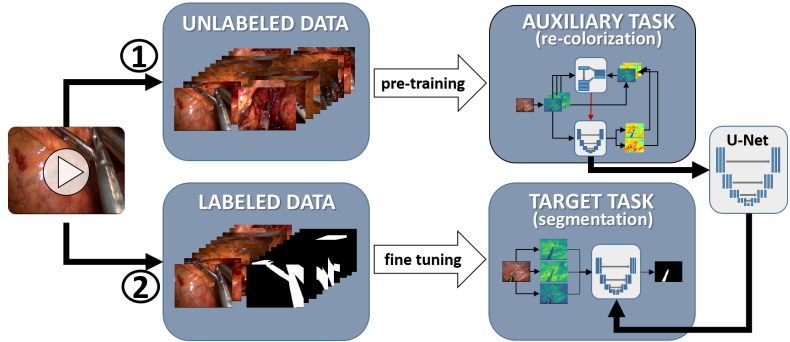

Fig. 2: Our approach applied to the specific task of instrument segmentation in endoscopic video data. A pre-training step leverages information available in unlabeled video data from the target domain. In this study, GAN-based re-colorization of video data (i.e. mapping the l-channel to the a, b - channel) was chosen as auxiliary task. The labeled training dataset is then used for fine-tuning the net according to the target application (i.e. segmentation).

*2.2.1 Auxiliary task: re-colorization*

The complete architecture of our pre-training is illustrated in Fig. 3. To train re-colorization, we first transform all images into the *CIE 1976 L\*a\*b\* Color space*. The axes $(L,a,b)$ of the color space are defined by the luminescence (L-channel), the color gradient from green to red (a-channel) and the the color gradient from from blue to yellow (b-channel). Using the L-channel as input, we train the network to predict the resulting a- and b-channels [30]. In this Context it is worth noting that quantitative automatic assessment of image similarity is challenging due to the lack of appropriate metrics. These often suffer from semantically valid changes which are imperceptible to humans (like slightly shifted pixels) [12]. To address this challenge, we adopt an additional Generative Adversarial Network (GAN) approach, as described in Larsen et al. [12]. GANs consist of two competing neural networks, a generator (see Par. Generator in sec. 2.2.1) and a discriminator (see Par. Discriminator in sec. 2.2.1) [8]. The discriminators role is to distinguish between real and fake images, while the generator tries to create fake images which fool the discriminator. This encourages the generator to produce better images and approximates the local data distribution [26]. Since most of the low level semantic information is already encoded in the L-channel, we only use the discriminator output. This is similar to a conditional GAN approaches like the pix2pix architecture [8,11].

*Generator*  As generator $G$, we use a U-Net [25] which, given the luminescence channel $I^l$ as input, predicts the corresponding a and b channels $\hat{I}^{a,b}$. In contrast to the original published U-Net architecture our blocks consists of two consecutive convolutional, followed by a batch normalization layer. Our final output layer is a tanh normalization. We train the generator U-Net to generate realistically re-colorized images with a loss function that is composed of three terms:

$$L_G = \gamma L_1 + \lambda L_2 + \varphi L_3 \tag{1}$$

with $\gamma, \lambda$ and $\varphi$ as weighting factors. The first loss term $L_1$ (see eq. 2) is the commonly used least squared GAN loss [18], defined by the output of the discriminator $D(\hat{I})$ for a fake image $\hat{I} = G(I^l)$ with the label $Y^D_{real}$. $L_1$ includes the output of the discriminator and forces so the production images which can fool the discriminators decision.

$$L_1(\hat{I}, Y) = \left\| D(\hat{I}_i) - Y^D_{real} \right\| i_2^2 \tag{2}$$

Due to a unbalanced distribution of color values and to improve the correct colorization of instruments or medical equipment, we extend the loss function of the generator $L_G$ with the term $L_2$ (eq. 3). The distribution of *ab* values of endoscopic images is strongly biased towards red, yellow, and black values, due to the appearance of background such as adipose tissue and blood, and the fact circular content area of most images is enclosed within a black background (see Fig. 4). Re-balancing is required to compensate for this and prevent the re-colorized images from being dominated by the most frequent values. Inspired by Zhang et al. [30], we obtain the empirical color distribution $\tilde{p}_c$ for each channel $c = \{a, b\}$ separately, with a quantization of the color space by a grid size of 1. The loss function $L_2$, as defined in eq. 3, ensures a high penalty for wrong values in image regions with rare values:

$$L_2(\hat{I}, I) = \frac{2}{h \cdot w} \sum_{c=a,b} \left[ \sum_{h,w} \left\| (\hat{I}^c_{h,w} - I^c_{h,w}) \cdot P^c_{h,w} \right\|_2^2 \right] \tag{3}$$

where, given by $\tilde{p}_c$, the weighting factor $P^c$ is the relative color frequency for each value in the input image. A quantized heatmap of the color distribution $\tilde{p}_c$ for the a and b-channel can be seen in Fig. 4. To prevent the learning from just rare values which would result in miss colorized images (for example a purple colouring), as an antagonist to $L_2$, we define $L_3$ to force the network to learn rare values and still be able to produce a valid colorization, similar to the original image.

$$L_3(\hat{I}, I) = \frac{1}{h \cdot w} \sum_{c=a,b} \left[ \sum_{h,w} \left\| (\hat{I}^c_{h,w} - I^c_{h,w}) \right\|_2^2 \right] \tag{4}$$

While the pre-training is only based on the l-channel, the target task makes additional use of the a and b channel. To this end, we added two zero initialized dummy input layers for the pre-training. After the pre-training, the network had learned to ignore the additional two empty channels, and during the target task training, these two layers were filled and the network learned to include them in its decision.

*Discriminator* We use an untrained ResNet18 [9] with the output of $D(I) \in [0,1]$ as discriminator $D(I)$. For training purposes, we show the network real images $I$ and re-colorized images $\hat{I}$. We use the mean squared error (MSE) with the labels $Y^D_{fake} = 0$ and $Y^D_{real} = 1$ [18] as loss function w.

$$L_D = MSE\left(D(I) - Y^D_{real}\right) + MSE\left(D(\hat{I}) - Y^D_{fake}\right) \tag{5}$$

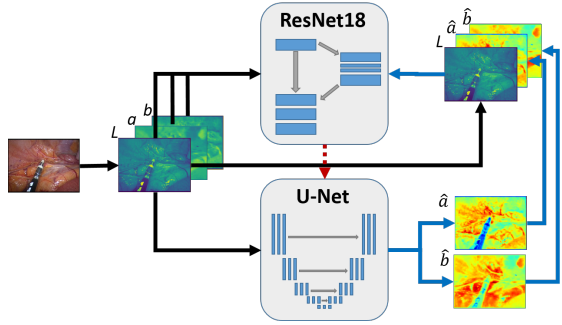

Fig. 3: Pre-training using self-supervised learning and a generative adversarial network (GAN) approach. First the image is transformed into the *LAB* color space. The luminescence layer l is fed into the generator $G(I^l)$ (U-Net) which is trained to generate the corresponding $\hat{a}$ and $\hat{b}$ channels. The discriminator $D$ (ResNet18) is trained to differentiate between real images $I = \{l, a, b\}$ from the target domain and fake images $\hat{I} = \{l, \hat{a}, \hat{b}\}$ produced by the generator.

*2.2.2 Target task: Instrument segmentation*

For target task training, we propose two variants of our method. The first one (**PAT**) does not require any additional labeled data while the second one (**PAT-Ext**) uses labeled data from a different but similar domain (in this instance: DS_MICCAI_L, see par. *Validation Data*.).

1. PAT: After pre-training our model as described in sec. 2.2.1, we use the U-Net that was pre-trained on the re-colorization task and fine-tune it for image segmentation. For this purpose, we adopt all pre-trained layers and only randomly initialize the last layer, which outputs the final segmentation $c \in \{INSTRUMENT, BACKGROUND\}$ for each pixel. We implement the cross entropy (see eq. 6) as loss function between the output and the groundtruth.

$$L_s(\hat{y}, y) = -y + log\left(\sum_{j \in c} e^{\hat{y}_j}\right) \quad (6)$$

2. PAT-Ext: We use additional available labeled data from a similar domain to extend the pre-training for the PAT-Ext model. To this end we extract the trained U-Net from the previously performed re-colorization task and re-initialize the last layer, similar to PAT. Following this, we re-train our U-Net on our additional available segmentation data as described above, before we finally fine-tune it on our sparsely labeled dataset.

## 2.3 Experiments

Based on endoscopic video data from two different sites (par. *Validation Data*), we (1) assessed the performance of our method as a function of the number of labeled images (par. *Effect of training data size*), (2) evaluate the effect of data augmentation on our method (par. *Effect of data augmentation*), (3) investigated the effect of different data domains used for pre-training with our method (par. *Effect of the unlabeled data domain*) and (4) compared our pre-training methods to related work using labeled data (par. *Comparison to other pre-training methods*).

*Validation Data*  We used the following datasets (L: labeled; UL: unlabeled) for validation purposes:

- **DS_COCO_L:** All 2,818 images of cats from the COCO dataset [14] and the corresponding segmentations. We chose cats as target class because unlike other classes of the COCO dataset, the corresponding images do not suffer from poor references or ambiguities [10] and have the target object in the foreground (similar to medical instruments in endoscopic data). Note, however, that the color distribution of **DS_COCO_L** is comparable to that of the whole COCO dataset.
- **DS_COCO_UL:** 20k natural images from the COCO dataset [14]. Comprises all 2,818 cat images and an additional 16,692 images selected randomly from the remaining 91 classes.
- **DS_MICCAI_L:** 2,400 endoscopic images with the corresponding instrument segmentations as used by the robotic instrument segmentation challenge[3] that was part of the MICCAI endoscopic vision challenge 2017. The sets training/testing images are disjunct and already predefined by the challenge.
- **DS_MICCAI_VID:** 21 unpublished endoscopic videos, which the images of DS_MICCAI_L were extracted from.
- **DS_HD_UL:** 30 endoscopic videos used in the surgical workflow challenge[4] that was part of the MICCAI endoscopic vision challenge 2017.
- **DS_HD_L [3]:** 809 annotated images from 6 surgeries with the corresponding binary instrument segmentations. The images were extracted from the DS_HD_UL dataset. We split our data into a training set of 413 (three surgeries), validation set of 119 (one surgery), and test set of 277 images (two surgeries). The sets of videos corresponding to testing images and training/validatation images are disjunct and randomly chosen.

*Training hyperparameters*  Unless otherwise stated in the method section, all networks ( ResNet18, U-Net) that were trained for the experiments have the same architecture as described in their original publications. [9, 25]. In this manuscript, we only mention important deviations from the original implementations. Hyperparameters were optimized with 80% of the training data using a fixed validation data set of $\frac{1}{5}$ of the training data. We used a preliminary hyperparameter space search and kept the parameters fixed for all consecutive experiments. For the re-colorization task we used the adam optimizer with a batch size of 12 and a learning rate of 0.0005 for the generator and 0.002 for the discriminator. Following visual exploration of our loss function on validation data, we stopped the pre-training after 20 epochs. For our experiments, we set the parameter $\gamma, \lambda$ and $\varphi$ such that the scale of $L_1, L_2$ and $L_3$ are equally ranked. For every unlabeled dataset (dataset name ending with UL, see par. *Validation Data*) we achieved a color distribution $\tilde{p}$ which was used later on for the re-colorization training on the corresponding dataset. The training of the the instrument segmentation was done with the adam optimizer and a learning rate of 0.0005. We used a scheduler that reduced the learning rate by a factor of 0.1 after a loss plateau lasting for 10 epochs. The batch size was 6. We stopped the training routine after 150 epochs.

*Effect of training data size*  To investigate the performance of our PAT method and its variant PAT-Ext (cf. sec. 2.2.2) as a function of the number of labeled training images, we used the Dice Similarity Coefficient (DCS) and the Intersection over Union (IoU) as target metrics. Based on DS_HD_L, we generated training datasets of size $k \cdot N$ with $k \in \{1, \frac{1}{2}, \frac{1}{4}, \frac{1}{8}, \frac{1}{16}\}$ and $N = 413$ denoting the total number of labeled training images available. For each $k$, five randomly selected disjunct subsets of training and validation data were generated (if possible). For PAT-Ext, we used the labeled dataset DS_MICCAI_L for model refinement. Testing for all our experiments was done on the complete DS_HD_L test dataset.

---

[3] https://endovissub2017-roboticinstrumentsegmentation.grand-challenge.org/
[4] https://endovissub2017-workflow.grand-challenge.org/

*Effect of data augmentation* Data augmentation is commonly used, especially if just a small amount of training data is available [7]. To investigate whether the method proposed complement the benefits of data augmentation, we repeated the experiments described in paragraph *Effect of training data size* with the original training data complemented via data augmentation. We used mirroring, rotation (90°, 180°, 270°) and adding of Gaussian noise (20%). All transformations were randomly applied by a 50% chance.

*Effect of the unlabeled data domain* We compared the performance of the PAT method when instantiated with different datasets, which were from the target domain (DS_HD_UL), a similar medical domain (DS_MICCAI_UL) and a non-medical domain (DS_COCO_UL), to examine the effect of the different domains. After pre-training all models were fine-tuned on DS_HD_L with the varying amount of data and tested as described in par. *Effect of data augmentation*.

*Comparison with other pre-training methods* We implemented two commonly applied state-of-the-art pre-training methods with labeled data:

1. **SOA (non-medical):** We performed a segmentation pre-training with the non-medical dataset (DS_COCO_L) and then fine-tuned the net with data from DS_HD_L, as described in par. *Effect of training data size*.
2. **SOA (medical):** Similarly to SOA(non-medical), we performed pre-training with a medical dataset representing a similar domain (DS_MICCAI_L).

These methods were compared with our approach using unlabeled data (PAT) only, while using unlabeled and labeled data for pre-training (PAT-Ext). For a description of PAT and PAT-Ext see sec. 2.2.2).

## 3 Results

*Performance of re-colorization* According to our experiments, re-colorization of images using the proposed GAN-based approach work produces realistically looking images when trained on medical data (see Fig. 4). In contrast, training on natural images as provided by the DS_COCO_UL dataset results in re-colored images that do not resemble the endoscopic images encountered in practice. Quantitative assessment of the method is (indirectly) provided in the following paragraphs when investigating the effects of the pre-training method on the segmentation method.

*Effect of training data size* In order to investigate whether the results of our methods are statistically significantly better than those of the baseline method, we calculated the mean DSC across all 5 splits for each test image of a fraction separately and performed an arc-sine transformation to obtain normally distributed data. Subsequently, we fitted linear mixed models [19] on each fraction $\frac{1}{20}$ to $\frac{1}{2}$ with the training method and the image as fixed and random effect, respectively. Resulting p-values from comparisons of baseline, PAT and PAT-Ext were adjusted by Dunnett's test and Bonferroni-Holm correction for multiple testing with fractions and across fractions, respectively. The analysis shows that both PAT and PAT-Ext are significantly better ($\alpha = 0.05$) than the baseline within each fraction between $\frac{1}{20}$ and $\frac{1}{3}$ with p-values $< 0.001$, except for the fraction $\frac{1}{20}$ in the PAT method with a p-value of 0.01. The p-values for fraction $\frac{1}{2}$ are 0.89 and 0.53 for PAT-Ext and the baseline. The median difference of DSC between the compared methods for fraction $\frac{1}{20}$ to $\frac{1}{3}$ lies within the range $[0.04, 0.06]$ for PAT and for PAT-Ext within $[0.04, 0.13]$. For fraction $\frac{1}{2}$ differences in performance are negligible. The performance of our method compared to the baseline method (no pre-training) is shown in Fig. 5a. In all experiments, our pre-training method based solely on unlabeled data (PAT) clearly boosted the performance of the segmentation method. When using $\frac{1}{16}$th of the training images (i.e. 25 images) we obtained a higher median performance than the baseline method trained with $\frac{1}{8}$ of the data. This corresponds to a decrease in the manual annotation effort of over than 60%. Analogously, we reduced the laboring effort by more than 50% and by around 25% when training on $\frac{1}{8}$th and $\frac{1}{4}$th of the training data. Even better results can be obtained when combining our pre-training for unlabeled data with pre-training using labeled data for the target task from a similar domain (PAT-Ext). Tab. 1 provides descriptive statistics for both target metrics (DSC and IoU) when using $\frac{1}{16}$th and $\frac{1}{8}$th of the training set images.

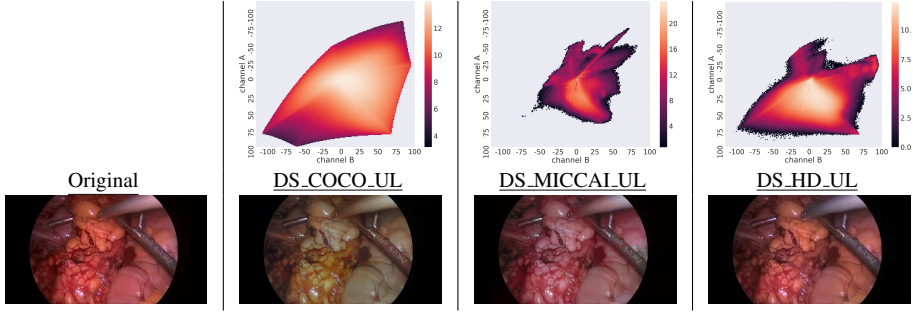

Fig. 4: Log color distributions of the a and b channels in the Lab colorspace of the corresponding dataset. Based on this distribution, the re-coloring has been done and leaded to the different color reconstructions as not all colors are equally represented. Red and yellow values occur more frequently in endoscopic videos than outside of the endoscopic context (DS_COCO_UL)

*Effect of data augmentation* As shown in Fig. 5b, data augmentation leads to a similar increase in performance for both the baseline method and the methods proposed. However, we could explore better results by combining our pre-training for unlabeled data with a pre-training using labeled data from the target task from a similar domain (PAT-Ext), than by just applying data augmentation, especially in fractions with less than $\frac{1}{6}$ of the training data. The best performance was achieved by extending the training of PAT with data augmentation during the fine-tuning step (PAT-Ext augmented).

*Effect of pre-training domain* Fig. 6 shows the performance of our method for different domains used in the pre-training process. Regardless of the number of labeled training images used for model fine-tuning, the best results are achieved when the pre-training is performed on the target domain. Using non-medical data decreases accuracy but still provides better performance than the baseline (no pre-training).

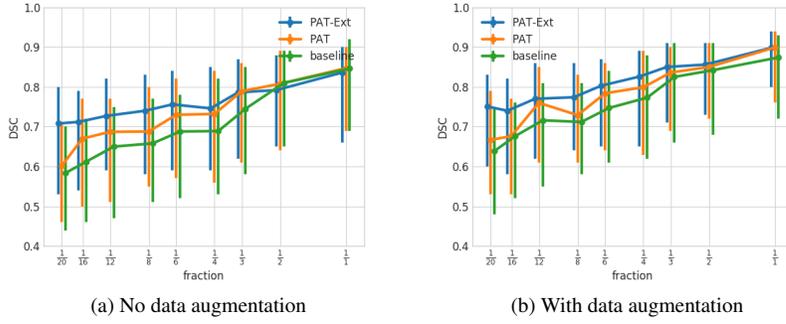

(a) No data augmentation

(b) With data augmentation

Fig. 5: Median Dice Similarity Coefficient (DSC) and the Interquartile Range (IQR) as a function of training data size as described in par. *Effect of training data size*. Our method clearly outperforms the baseline method without pre-training and is even better than data augmentation by very small numbers of available training data.

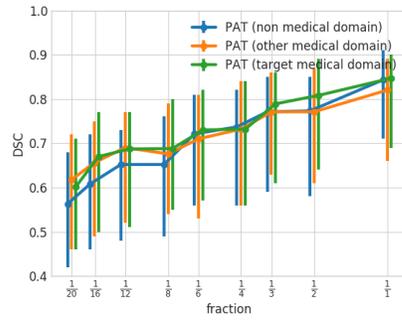

Fig. 6: Effect of pre-training domain. For small training datasets, medical images yield better results than non-medical images.

*Comparison and combination with other pre-training methods* We extended our comparison for the performance of our method to state-of-the-art pre-training methods that rely on labeled data. Our method outperformed the state-of-the-art approach of applying pre-trained nets from a non-medical domain (SOA (non-medical)) that have been trained with labeled images for the target task. However, training on labeled images of a similar domain (SOA (medical)) generally yielded better results than pre-training exclusively on unlabeled data (PAT). The best results were achieved when combining state-of-the-art pre-training on medical data with our approach (PAT-Ext). Detailed results are listed in table 1. The more training images used for fine-tuning, the more similar the results of all methods are.

## 4 Discussion

While approaches to semi-supervised learning, which typically handle unlabeled and labeled data from the same data distribution simultaneously, are increasingly common in the field of Medical Image Computing (MIC) [2], we are, to our knowledge, the first to investigate

Exploiting the potential of unlabeled endoscopic video data with self-supervised learning        11Table 1: Descriptive statistics for the target metrics DSC and IoU when using $\frac{1}{16}$th, $\frac{1}{8}$th of the training set images (i.e. 102, 50 and 25 labeled images for final model refinement). Two state-of-the-art (SOA) methods (see par. 2.3) are compared to three variants of our method (see sec. 2.2.2). The mean values are shown along with the improvement in % compared to the baseline method (no pre-training).

dataset fraction: $\frac{1}{16}$

| Training type | DSC Mean | DSC Median | DSC IQR | IoU Mean | IoU Median | IoU IQR |
|---|---|---|---|---|---|---|
| baseline | 0.57 (-) | 0.61 | [ 0.46 0.72] | 0.44 (0%) | 0.47 | [ 0.31 0.57] |
| SOA (non-medical) | 0.57 (-1%) | 0.62 | [ 0.43 0.73] | 0.44 (0%) | 0.47 | [ 0.30 0.59] |
| SOA (medical) | 0.59 (3%) | 0.65 | [ 0.47 0.75] | 0.46 (5%) | 0.50 | [ 0.33 0.61] |
| PAT (target medical) | 0.61 (7%) | 0.67 | [ 0.50 0.77] | 0.49 (10%) | 0.52 | [ 0.36 0.63] |
| PAT (other medical) | 0.60 (5%) | 0.66 | [ 0.49 0.75] | 0.47 (7%) | 0.51 | [ 0.34 0.61] |
| PAT-Ext | 0.65 (13%) | 0.71 | [ 0.54 0.79] | 0.53 (19%) | 0.57 | [ 0.38 0.66] |

dataset fraction: $\frac{1}{8}$

| Training type | DSC Mean | DSC Median | DSC IQR | IoU Mean | IoU Median | IoU IQR |
|---|---|---|---|---|---|---|
| baseline | 0.62 (0%) | 0.66 | [ 0.51 0.77] | 0.49 (0%) | 0.51 | [ 0.36 0.64] |
| SOA (non-medical) | 0.63 (1%) | 0.67 | [ 0.52 0.78] | 0.50 (2%) | 0.52 | [ 0.38 0.65] |
| SOA (medical) | 0.66 (6%) | 0.71 | [ 0.53 0.82] | 0.54 (10%) | 0.59 | [ 0.39 0.71] |
| PAT (target medical) | 0.65 (5%) | 0.69 | [ 0.55 0.80 ] | 0.53 (6%) | 0.54 | [ 0.41 0.67] |
| PAT (other medical) | 0.65 (5%) | 0.68 | [ 0.54 0.79] | 0.52 (5%) | 0.53 | [ 0.40 0.67] |
| PAT-Ext | 0.68 (10%) | 0.74 | [ 0.58 0.83] | 0.56 (14%) | 0.60 | [ 0.42 0.71] |

the concept of self-supervised learning to reduce manual labeling effort in medical image segmentation. In contrast to state-of-the-art pre-training methods [24, 27, 28, 32], we initialized our model on the target domain using only unlabeled data rather than on a different domain with labeled data. This is achieved with an auxiliary task that can be assumed to learn a representation of the target domain that is well-suited for the target task. According to the experiments in this study, our approach is suitable for leveraging information in unlabeled endoscopic video data to reduce the amount of labeled training required. Our method not only outperformed the baseline method without pre-training by a large margin, but also yielded better results than the state-of-the-art pre-training method requiring labeled data. This is particularly apparent in small sets of labeled data. The related literature on pre-training with self-supervised learning is very recent (with some of it being produced in parallel to our work) and is mainly proposed by the computer vision community. Analysis of various auxiliary tasks (inpainting [22], re-colorization [13, 30, 31], classification [20, 30], re-ordering [20] and prediction [1]) for multiple applications suggests that re-colorization is the most promising approach for a number of applications. The closest work to ours was recently authored by Bodenstedt et al. [4], who introduced an auxiliary task that estimates the order of appearance of two video frames in order to pre-train a CNN for surgical phase recognition. To our knowledge, however, using re-colorization as auxiliary task has not yet been investigated in the field of medical image analysis. Our results suggest that the method proposed complements the benefits gained from data augmentation. If we compare the improvement in performance resulting from the two complementary methods, it can be concluded that our method is particularly well-suited to situations where only little training data (up to $\frac{1}{6}$ of the training data) is available. In these cases, the benefits of PAT-Ext pre-training are greater than those for data augmentation. It should be pointed out that the goal of this work was not to optimize the performance of an algorithm for a specific application. Instead, our aim was to explore ways to make optimal use of available data sources. An additional increase in accuracy could, for example, be gained by optimizing the weights in our loss function. However, an interesting side-effect is that our method achieves state-of-the-art performance on the most recent MICCAI endoscopic vision dataset for instrument segmentation (Fig. 1), even without data augmentation. In contrast, absolute performance is much worse on our own dataset. We attribute this to the comparatively low variability of the MICCAI images as well as the challenging nature of our own images. The auxiliary task chosen in this paper (GAN-based re-colorization) appears to be a very good match for the target task of medical instrument segmentation, as suggested by the experimental results. We are currently planning to test our method on further target tasks. Future work should be focused on finding optimal auxiliary tasks for a given target application.

In conclusion, we have developed a pre-training approach that makes optimal use of all the available data sources: both, public and non-public, in addition to labeled and unlabeled. As it can potentially be applied to a wide range of target tasks, the potential impact on the research community and possible clinical applications is high.

**Acknowledgements** We acknowledge the support of the European Research Council (ERC-2015-StG-37960). This work was support by Intuitive Surgical who providing us with the raw video data, from which the MIC-CAI2017 robotic challange data were extracted. We further acknowledge the support of the Federal Ministry of Economics and Energy (BMWi) and the German Aerospace Center (DLR) within the OP 4.1 projekt. Finally, we would like to thank Simon Kohl inspiring us to this paper.
**Conflict of Interest:** The authors declare that they have no conflict of interest.
**Ethical approval:** For this type of study formal consent is not required.
**Informed consent:** This article contains patient data from publically available datasets.